\documentclass[journal, 12pt, onecolumn]{IEEEtran}

\usepackage{adjustbox}
\usepackage{amssymb}
\usepackage{amsthm}
\usepackage[cmex10]{amsmath}
\usepackage{amsfonts}
\usepackage{algorithmic}
\usepackage[linesnumbered, ruled, vlined, onelanguage, ]{algorithm2e}
\usepackage{array}
\usepackage{authblk}
\usepackage[english]{babel}
\usepackage{blindtext}
\usepackage{booktabs}
\usepackage{cite}
\usepackage{colortbl}
\let\labelindent\relax
\usepackage{enumitem}  
\usepackage{eqparbox}
\usepackage[T1]{fontenc}
\usepackage[hang,flushmargin]{footmisc}
\usepackage{graphicx}
\usepackage{import}
\usepackage{latexsym}
\usepackage{lipsum}
\usepackage{listings}
\usepackage{hyperref}
\hypersetup{
  colorlinks   = true, %Colours links instead of ugly boxes
  urlcolor     = black, %Colour for external hyperlinks
  linkcolor    = black, %Colour of internal links
  citecolor   = black %Colour of citations
}
\usepackage{mdwmath}
\usepackage{mdwtab}
\usepackage{multicol}
\usepackage{multirow}
\usepackage{pifont} % http://ctan.org/pkg/pifont
\usepackage[tight,footnotesize]{subfigure}
\usepackage{stfloats}
\usepackage{supertabular}
\usepackage[dvipsnames,table]{xcolor}% enables use of color package
\usepackage{xparse}

\usepackage{silence}
\usepackage{caption}

\definecolor{Gray}{gray}{0.9}
\definecolor{LightCyan}{rgb}{0.88,1,1}
\usepackage[first=0,last=9]{lcg}

\newcommand{\bs}{\textbackslash}
\newcommand{\vsp}{\vspace{1em}}
\renewcommand{\indent}{\hspace{0.7cm}}
\rowcolors{2}{gray!25}{white}
\newcommand*\rot[1]{\rotatebox{90}{#1}}

\begin{document}

\title{\textcolor{white}{.}\\  Individual Explanations in Machine\\ Learning Models: A Survey for Practitioners}

\author{\IEEEauthorblockN{
\ \ \  \ \ \ Alfredo Carrillo \ \ \ \ \ \ \ \ \ \
Luis F. Cant\'u\ \ \ \ \ \ \ \ \  
Alejandro Noriega \ \  }

\IEEEauthorblockA{
alfcar9@prosperia.ai \ \ \ \ \
cantu@prosperia.ai   \ \ \ \ \
noriega@mit.edu}\\
}

\providecommand{\keywords}[1]{\textbf{\textit{Index terms---}} #1}

\maketitle
\thispagestyle{empty}

\begin{abstract}
In recent years, the use of sophisticated statistical models that influence decisions in domains of high societal relevance is on the rise.
Although these models can often bring substantial improvements in the accuracy and efficiency of organizations, many governments, institutions, and companies are reluctant to their adoption as their output is often difficult to explain in human-interpretable ways. Hence, these models are often regarded as black-boxes, in the sense that their internal mechanisms can be opaque to human audit. In real-world applications, particularly in domains where decisions can have a sensitive impact\textemdash e.g., criminal justice, estimating credit scores, insurance risk, health risks, etc.\textemdash model interpretability is desired. Recently, the academic literature has proposed a substantial amount of methods for providing interpretable explanations to machine learning models. This survey reviews the most relevant and novel methods that form the state-of-the-art for addressing the particular problem of explaining individual instances in machine learning. It seeks to provide a succinct review that can guide data science and machine learning practitioners in the search for appropriate methods to their problem domain.
\end{abstract}

\small
\hspace{0.75cm}
\normalsize

\tableofcontents

\section{Introduction}

\vsp
\subsection{Relevance of Interpretability for Machine Learning Models}

Machine learning (ML) has great potential for improving decisions, processes, policies, and products. In the past two decades, algorithmic decision-making has spread increasingly to many relevant societal contexts, including the provision of health services \cite{kleinberg2015prediction}, counter-terrorism \cite{pelzer2018policing}, criminal justice \cite{kleinberg2016inherent}, and assessment of credit and insurance risk \cite{wan2011prediction}, to name a few. 

\vsp
However, most ML methods are not inherently transparent, as they do not provide human-interpretable explanations of their outputs \cite{molnar2019interpretable}. Moreover, recently the most successful algorithms in difficult tasks, such as machine vision and natural language processing, are moving towards models of high complexity, such as deep neural networks, which has further increased the tension between accuracy and interpretability \cite{Goodfellow-et-al-2016}. The lack of interpretability often becomes a barrier to the adoption of these technologies in both public and private organizations. Conversely, if ML models' estimations are coupled with human-interpretable accounts of the statistical rationals behind their own opinions, then human decision-makers can more easily audit models, and better integrate models' estimations with human judgements and additional qualitative information.

\vsp
Within the paradigm of statistical learning and ML, the human role is not to explicitly define decision rules, but to formulate the prediction task, collect the ground truth, and allow a \textit{learning algorithm} to recognize patterns and find the best-performing decision rule through optimization. \iffalse Since humans have no intervention during the execution of the \textit{training} process, the track---or insight---into \textit{what} the machine is learning becomes unclear.\fi This process can generate very elaborate decision rules, particularly when using high-dimensional input, allowing to generate a high-degree of non-linear interactions across features. Hence, the resulting models are often \textit{complex}, and hard to boil down to a simple set of rules amenable to human intuition \cite{molnar2019interpretable}. 

\vsp
As stated by Molnar \cite{molnar2019interpretable}, there is not a rigorous definition for interpretability. Yet, we list two intuitive definitions below. Both definitions are similar, they seek a better understanding on how a model makes its decisions.

\begin{enumerate}[label=(\roman*)]
    \item \textit{Interpretability is the degree to which a
human can understand the cause of a decision} \cite{miller2019explanation}.
    \item \textit{Interpretability is the degree to which a human can consistently predict the model’s result} \cite{doshi2017towards}.
\end{enumerate} 

\vsp
Finally, several motivations exist for having an interpretability layer for ML models, including: a) being able to \textit{justify} predictions, b) being able to \textit{diagnose} vulnerabilities of models, to be able to c) \textit{improve} them \cite{adadi2018peeking}, and c) to \textit{understand} better our reality, as when algorithms discover relevant patterns or relations, we can build upon them to generate new knowledge.

\iffalse
\vsp
Another classical explainability problem is the phenomenon known as \textit{The Rashomon Effect}, which states that sometimes there are independent collections of reasons that can justify the model's prediction equally well. Finally, common ways for explaining are by: natural language, visualizations of learned
representations or models, and by example--contrasting an observation of interest against others. 
\fi

\subsection{Design of this Survey}

\textbf{Objective.} This survey aims to map and provide a taxonomy of state-of-the-art methods that provide individual explanations in ML. By doing so, it aims at aiding programmers and data scientists to find methods suitable for their context. In addition, it aims to contribute to research community by highlighting some of the open problems in the field. \iffalse Ultimately, this will accelerate the process for the common adoption of ML, from which the society as a whole will benefit substantially. \fi

\vsp
\textbf{Survey Methodology.} This survey focuses on relevant papers from IEEExplore, ACM Digital Library, and Google Scholar. We began searching for the following keywords: ``artificial intelligence'', ``black box'', ``explainability'', ``interpretability'', ``locally, ``transparency'', ``machine learning''. We also reviewed other surveys, and kept track of papers frequently cited. We considered papers from 2010 until 2020, with few relevant exceptions. The selection was based on conceptual contribution and number of citations in the academic literature.

\vsp
\textbf{Contents.} This survey is structured as follows. In Section \ref{taxonomies} we overview the main taxonomies used to classify interpretability methods. We delimit our scope to local methods, as they serve the purpose of giving individual explanations, and introduce two main approaches---Model-Agnostic and Model-Specific Methods. In Section \ref{model_agnostic}, we present and discuss explanations methods that follow a Model-Agnostic approach, while in Section \ref{model_specific} we focus on methods that follow a Model-Specific approach, providing further taxonomy for each. Section \ref{conclusion} concludes.

\section{Taxonomies of Explainability Methods}
\label{taxonomies}

\subsection{Scope: Local vs Global} 

In this survey, we focus on methods that give explanations for individual predictions. These methods are called (i) \textit{Local Methods}. In contrast, (ii) \textit{Global methods} describe the behavior of the model as a whole.

\begin{enumerate}[label=(\roman*)]
    \vsp
    \item \textbf{Local methods.} These aim to understand why did the model make a certain prediction for a given instance or group of nearby instances. Perhaps, the model is too complex, but by delimiting to a neighborhood, it is possible to describe the model locally with a set of simple rules. In the best case, the prediction will be determined locally by only a low number of interactions from the features, with a monotonous or even linear effect \cite{molnar2019interpretable}.
    
    \vsp
    \item \textbf{Global Methods.} These aim at understanding the behavior of the model as a whole \cite{doshi2017towards}. Global methods answer to the question: how do different elements of the model or input features affect predictions? However, aiming at global interpretability is a difficult challenge, mainly in highly complex models. Therefore, it is common for global methods to lower the bar and attempt to understand the model up to a modular level \iffalse Lastly, some authors propose use the strength of both approaches\fi \cite{guidotti2018local, linsley2018global, seo2017interpretable}.
\end{enumerate}

\vsp
Needless to say, these are not in conflict. It is possible to approach to a global understanding of the model by aggregating local knowledge. First, by understanding individual instances, moving on to a cluster or group of instances. From there, to a modular level, and finally to a global level. The converse is straightforward \cite{ribeiro2016should}. If we can interpret a model globally, we can explain a single prediction.

\subsection{Generality: Model-Agnostic vs Model-Specific}
A second common approach to classify explainability methods is by Model-Generality. In this taxonomy, a method can be (i) \textit{Model-Agnostic} or (ii) \textit{Model-Specific}. 

\begin{enumerate}[label=(\roman*)]
    \vsp
    \item \textbf{Model-Agnostic} methods are independent from the model used. Therefore, these post-hoc methods can be applied to any model. They cannot have access to the internal structure of the model, only to inputs and outputs.
    
    \vsp
    \item \textbf{Model-Specific} interpretations tools are limited to a specific model, as they make use of this internal structure. For example, the weights in a linear model, or the cut-offs in a decision tree. 
\end{enumerate}

\subsection{Other Taxonomies} 
Finally, another common taxonomy for classifying explanations methods is by Interpretability Degree: (i) \textit{Post-Hoc} or (iii) \textit{Intrinsically Interpretable}. 

\begin{enumerate}[label=(\roman*)]

\vsp
\item \textbf{Post-Hoc} methods, which apply interpretable methods to the model after training.

\vsp
\item \textbf{Intrinsically Interpretable methods}, which are a particular case of ad-hoc methods, in which the model itself has a high degree of interpretability---as opposed to highly complex models. Two of these intrinsically interpretable methods are linear regressions  and classification trees.
\end{enumerate}

\subsection{Reviewed Methods} 

Similar to the table shown in \cite{adadi2018peeking}, in Table \ref{methods}, we enlist the methods that are reviewed in this survey. The table is divided in two parts. In the upper one, we have the Model-Agnostic Methods, described in Section \ref{model_agnostic}, whereas the lower part, discusses the Model-Specific Methods which appear in section \ref{model_specific}. In the \textit{Model} column, the type of model is reported. 
The \textit{Scope} column we have either a Local (L) method, or both, Global and Local (G\bs L), but not global methods only. (G\bs L) means the method serves both purposes. The \textit{Year} is for when the article was published. The fourth column contains the reference to the article and the fifth column is the number of citations (NC) according to Google Scholar during June 2020. The ordering of the methods is somewhat arbitrary, although we try to cluster, the best as we could, the methods that were related to one another. The thin horizontal lines denote the sub-classification within each of the two General approaches.

\begin{table}[ht!]
\centering
\caption{Survey's discussed methods}
\resizebox{0.8\columnwidth}{!}{
\label{methods}
\begin{tabular}{c|l|c|c|c|c|c|}
   \multicolumn{1}{l}{}  & \multicolumn{1}{l}{\textbf{Model-Agnostic Methods}} & \multicolumn{5}{c}{} \\ \cmidrule{2-7}
    & Method Name                        & Model & Scope  & Year & Article & NC    \\ \cmidrule{2-7}
    \multirow{11}{*}{\rot{\textbf{Perturbation-Based}}}
    & PDP                                & A     & G\bs L & 2001 & \cite{friedman2001greedy}         & 10,353 \\
    & ICE                                & A     & G\bs L & 2015 & \cite{goldstein2015peeking}       & 244    \\
    & ALE                                & A     & G\bs L & 2016 & \cite{apley2016visualizing}       & 75     \\
    & Shapley Values (SHAP)              & A     & L      & 2017 & \cite{lundberg2017unified}        & 1,212  \\
    & LOCO                               & A     & G\bs L & 2018 & \cite{lei2018distribution}        & 103    \\
    & Decomposition of pred.             & A     & L      & 2008 & \cite{robnik2008explaining}       & 195    \\
    & Feature Importance                 & A     & G\bs L & 2018 & \cite{casalicchio2018visualizing} & 24     \\
    & Sensitive Analysis                 & A     & G\bs L & 2013 & \cite{cortez2011opening}          & 225    \\
    & LIME                               & A     & L      & 2016 & \cite{ribeiro2016model}           & 3,236  \\
    & Explanations Vectors               & A     & L      & 2010 & \cite{baehrens2010explain}        & 503    \\
    & Anchors                            & A     & L      & 2018 & \cite{ribeiro2018anchors}         & 329    \\ \cmidrule{2-7}
    \multirow{3}{*}{\rot{\textbf{Contrast}}}
    & Counterfactuals                    & A     & L      & 2017 & \cite{wachter2017counterfactual} & 363    \\
    & Prototype and Criticism            & A     & G\bs L & 2016 & \cite{kim2016examples}           & 182    \\  
    & Justified Counterfactual           & A     & L      & 2019 & \cite{laugel2019dangers}         & 15     \\
    \cmidrule{2-7}
    \multicolumn{5}{l}{}\\

    \multicolumn{1}{l}{} &  \multicolumn{1}{l}{\textbf{Model-Specific Methods}} & \multicolumn{5}{c}{}\\ \cmidrule{2-7}
    \multirow{5}{*}{\rot{\textbf{Vision CN}}}
    & Masks                       & CN & L     & 2017 & \cite{fong2017interpretable}     & 393   \\
    & Real Time Saliency Map      & CN & L     & 2017 & \cite{dabkowski2017real}         & 151   \\
    & SmoothGrad                  & CN & L     & 2017 & \cite{smilkov2017smoothgrad}     & 326   \\
    & Layer-wise Relevant         & CN & L     & 2015 & \cite{bach2015pixel}             & 1,001 \\
    & Heat Maps                   & CN & L     & 2014 & \cite{zeiler2014visualizing }    & 9,516 \\ \cmidrule{2-7}
    \multirow{6}{*}{\rot{\textbf{General NN}}}
    & Differentiable Models       & NN & L     & 2017 & \cite{ross2017right}             & 140   \\
    & DeepLIFT                    & NN & L     & 2016 & \cite{shrikumar2016not}          & 157   \\
    & Taylor decomposition        & NN & L     & 2017 & \cite{montavon2017explaining}    & 432   \\
    & Integrated Gradients        & NN & L     & 2017 & \cite{sundararajan2017axiomatic} & 696   \\
    & I-GOS                       & NN & L     & 2019 & \cite{qi2019visualizing}         & 8     \\
    & Grad-cam                    & NN & L     & 2017 & \cite{selvaraju2017grad}         & 2,160 \\ \cmidrule{2-7}
    \multirow{1}{*}{\rot{\textbf{DTM}}}
    & TreeExplainer               & DT & L     & 2020 & \cite{lundberg2020local}         & 176   \\ \cmidrule{2-7}
    \multicolumn{6}{l}{\textit{\ \ \ \ A: Agnostic Model, NN: Neural Network, CN: Convolutional Network,}}\\ 
    \multicolumn{6}{l}{\textit{\ \ \ \ DT: Decision Tree, G: Global, L: Local, Global and Local (G\bs L)}}\\
    \multicolumn{6}{l}{\textit{\ \ \ \ DTM: Decision Trees Methods}}\\
\end{tabular}
}
\end{table}

\section{Model-Agnostic Methods}
\label{model_agnostic}
In this section it is discussed Model-Agnostic methods, one of the two taxonomies from the Method-Generability approach. The Model-Agnostic Methods do not depend on the specific model used; instead, these are methods that separate the explanation from the machine learning model \cite{molnar2019interpretable}. This technique consists of adding a new layer between the black-box and the human for interpretability purposes. These methods cannot access the internal parameters of the model, and they analyze the model using a sample of inputs and corresponding outputs to assess how it internally works. 

\vsp
Before discussing their advantages and disadvantages, we expose first the general desirable properties for Interpretability Methods, so that we can relate them with the Model-Agnostic Methods.

\begin{enumerate}[label=(\roman*)]
    \vsp
    \item \textbf{Model flexibility}: It refers to the ability of a method to work with any kind of model. 
    
    \vsp
    \item \textbf{Explanation flexibility}: It refers to the fact that there are several forms to present explanations. Such as natural language, visualizations of learned representations or models, and by example--contrasting an observation of interest against others. 
    
    \vsp
    \item \textbf{Representation flexibility}: It refers to the capacity of the method of not necessarily explaining via the input features. For example, a text classifier can receive sentences as input, and the explanation can be by individual words \cite{ribeiro2016model}. 
\end{enumerate}

\vsp
Two \textit{advantages} of these types of models are: 
\begin{enumerate}[label=(\roman*)]
    \vsp
    \item Model flexibility: Independently of which model is used to solve a task, any of these explanations methods can be used because they only use inputs and outputs to provide interpretability of the model, instead of using the particular internal mechanisms of each model.
    
    \vsp
    \item A model-agnostic method is useful for comparing different models by simply applying the same interpretability methodology to each.
\end{enumerate}

\vsp
Whereas, two \textit{disadvantages} are:
\begin{enumerate}[label=(\roman*)]
    \vsp
    \item Executing these types of methods can take a long lapse of time if the models are highly complex, or if the instances have a lot of features.
    
    \vsp
    \item The Model-Agnostic methods can suffer from sampling variability \cite{lundberg2020local}. For explaining models, these methods require a sample of inputs and outputs, and it is via this sample that the method tries to interpret the model. But because the sample is only a small fraction of the entire population of possible real-observable instances, the explanations vary---hopefully not too much---according to the sample selected.
\end{enumerate}

\vsp
\textbf{Model-Agnostic Method Approaches}:
\begin{itemize}
    \vsp
    \item \textbf{Perturbation Approach.} Almost all the methods reviewed in this section follow this approach. The explanations are given as feature-wise effects, or impacts on the output, according to the particular input values. For estimating each effect, a perturbation approach is commonly followed. A sample of artificial instances is generated via observed instances in the data set by distorting--or perturbing--slightly a few numbers of features while leaving the rest unchanged. The specific way to generate the sample varies across methods. In this particular context of individual explanations, the perturbations are centered on the instance of interest. These methods are reviewed in Section \ref{perturbation}.

    \vsp
    \item \textbf{Contrastive Approach} It explains by contrasting a focal instance against a close by reference group. Some methods use the same label prediction for the reference group and focal instance, while others use a different one. In the former fashion, the explanation consists on demonstrating that the focal instance belongs to a cluster in which similar predictions are made. Whereas for the latter, the explanations are counterfactual. The seek for the smallest changes of the focal instance such that the prediction label changes. These are simple explanations and useful to offer a recourse for users who obtained unfavorable decisions \cite{garg2019counterfactual}. These methods are reviewed in Section \ref{contrastive}.
\end{itemize}

\subsection{Perturbation Approach}
\label{perturbation}

A common way to explain an individual instance is via the effects that each feature's value contributed in the final explanation. In all of the following methods, this approach is followed. The corresponding article, with its year, is stated at the beginning of each method description.

\vsp
\textbf{Partial Dependence Plots (PDP).} A popular method introduced by Friedman \cite{friedman2001greedy} in 2001 is the PDPs. By using a plot, we explain by the marginal effect on the outcome across the feature's domain. We plot the partial dependence feature's function, which must be estimated by Monte Carlo method. The function's output is interpreted as the average outcome value if all observations had that feature value. It is also possible to plot the combined effect of a pair of features. The PDP method assumes the features are not correlated, but when this fails--situation extremely common in real-world data sets--the average is done over instances that are unlikely to occur. Therefore, the effect of the feature can be poorly estimated. The second problem is that it estimates the average outcome but not its variance. Thus, an effect can camouflage if it gets canceled out by a combination of observations positively and negatively correlated \cite{molnar2019interpretable}. As a final note, the PDPs is a global method, but one could use it to understand a single instance as follows. First, generate a PDP for every feature, and with the observation of interest, find the corresponding effect for each feature value. The upcoming three methods try to patch some of the inherent problems mentioned for PDPs, which is another reason why reviewing it, despite not being purely local. 

\vsp
\textbf{Individual Conditional Expectation (ICE).} In 2015, Goldstein et al. \cite{goldstein2015peeking} proposed another graphic method for interpretability. An ICE plot visualizes the effect of a feature in a graph, but plots the instances independently as lines. Instead of averaging and plotting a single line, here we have a line for each observation. The average of these multiple lines is a PDP. To generate one for a given instance in a particular feature, we do the following. First, we create a grid of values across the domain for that feature of interest. While leaving the rest of the features constant, we obtain a prediction. Finally, we use the sequence of outputs to produce a line. The ICE can help visualize better than the PDPs the effects of interactions. If there are no interactions, the lines will look very similar to one another. A considerable advantage with respect to PDPs is that now effects cannot longer be canceled out. This method still suffers if there is a correlation across features. An individual instance can be analyzed by examining how the outcome would change by perturbing each feature. Also, we can make comparisons of these effects across features. 

\vsp
\textbf{M-Plots.} The graphic method Marginal-Plots (M-Plots) was proposed to plot the features' true effects in a correlated dataset. It invited us to generate the plot as in the PDPs, with artificial observations, but in each iteration, it restricts the replacement of values to those that appear in a small neighborhood instead of using the whole feature's domain. This adjustment prevents the averaging of instances that are not found in reality. Unfortunately, the solution is partial because each feature's effect is mixed with the effects of its corresponding correlated features. 

\vsp
\textbf{Accumulated Local effects (ALE).} In 2016, Apley et al. proposed  \cite{apley2016visualizing} a graphic method called ALE plots for a scenario with correlation across features. It is an alternative technique that mitigates the problem of M-Plots. It consists of plotting the differences in predictions, instead of averages. The use of differences annuls the effect of other correlated features because we add and subtract the interaction's effects \cite{molnar2019interpretable}. The interpretation of an ALE plot centered at zero is as follows. Conditioning on a particular value, it is the relative effect on the output when changing the feature's value. For an individual explanation, we could use the same methodology described for the PDPs.

\vsp
\textbf{Shapley Values (SHAP).} From early 2001, an approach using Game Theory has been developed for individual explanations \cite{lipovetsky2001analysis}, where the idea is to use Shapley Values for interpretability. These values portray how to fairly distribute the contribution, or effect, for a prediction among its feature values \cite{molnar2019interpretable}. We break down the output as a sum of positive and negative effects, corresponding each to a feature's value. The Shapley value relative to a feature's value is defined as the average of all the marginal contributions to all possible coalitions. In detail, suppose we want to estimate the Shapley value for feature1-value1. A coalition is a subset of the set $\{$feature2-value2, feature3-value3, $\dots$, feature$p$-value$p\}$. A marginal contribution is the difference of two predictions from the same coalition, one with the feature1$=$value1, and another with feature1$\neq$value1. All but one coalition has missing values for prediction. We replace values with corresponding feature values from instances drawn randomly. Naturally, as the number of features increases, it becomes computationally costly. In 2010, Erik Strumbelj and Igor Kononenko \cite{strumbelj2010efficient} presented a method for classification models, which uses Shaply Sampling values the estimation. The novel, and popular article by Lundberg \cite{lundberg2018explainable}, presented an improvement in the methodology to calculate these values.

\vsp
\textbf{LOCO.} In 2018, Lei et al. \cite{lei2018distribution} proposed another technique for generating local-surrogate models called Leave-one-covariate-out (LOCO). The interpretability is based on evaluating the importance of each feature. This method is similar to the one developed by Breiman \cite{breiman2001random} for random forests. The idea is to measure the error in prediction error by artificially deleting one covariate and comparing it against using it. By a simulation process, the prediction error is estimated using confidence intervals for each possible value of the feature's domain. If the range of the interval does not share an intersection with zero, then the feature around that neighborhood helps to significantly drop the output's error.

\vsp
\textbf{Decomposition of predictor.} Robnik-Sikonja and Kononenko \cite{robnik2008explaining} proposed in 2008 to explain individual instances by decomposing a model's predictions on feature-wise effects. To explain a feature's relevance, they perturb its value and calculate its corresponding output. Then they analyze the difference between that output and the one corresponding to the instance without perturbations. This difference depicts the relevance that feature value is with respect to the output. They follow a univariate approach since they perturb one feature at a time while keeping the rest constant. This method does not work well when there is a high correlation across features since its common that in real-world scenarios, there are features pairs that are highly correlated, and with this approach, this fact is overlooked. Thus, the artificial instances generated are hard to observe in real-work scenarios.

\vsp
\textbf{Feature Importance.} In 2018, Casalicchio et al. \cite{casalicchio2018visualizing}, proposed a visualization method based on feature importance. The method is related to PDP and ICE plots, but visualize the expected conditional feature \textit{importance} instead of the expected conditional \textit{prediction}. Furthermore, the plots they produce are called individual conditional importance (ICI), analogous to the ICE but for importances. They demonstrate that by averaging ICI curves across observations yields to a curve that, if integrated with respect to the distribution, a measure that can be interpreted as the global feature importance can be obtained. Finally, they adapt the SHAP method to calculate \textit{SHAP feature's importance}. 

\vsp
\textbf{Sensitive Analysis.} In 2011, Cortez and Embrechts \cite{cortez2011opening} followed a visualization approach, where they observe how the input changes as they make perturbations over the features. Past works have used the same approach but perturbing at most two features at a time. In this work, they generalized it for an arbitrary number of features, although it has a more computational cost. For visualization purposes, they used the Variable Effect Characteristic curve, which plots the average impact of a particular input across its domain. The sensitivity measures are used to evaluate the change in the prediction as we vary the features. The same authors \cite{cortez2013using}, in 2013, published a second article where they extended their investigation. They proposed a new sensibility measure, as well as useful visualization plots for the results. We can use these methods locally and globally. For a particular instance, we can visualize the feature's importance with the aforementioned methodology.

\vsp
\textbf{LIME.} In 2016, Ribeiro et al. \cite{ribeiro2016should}, in their paper Local Interpretable model-agnostic Explanations (LIME), proposed using local-surrogate models. It is a post-hoc method that is trained on a locality of the black-box to make individual explanations. Taking an instance of interest, it examines counterfactuals that randomly perturb features in a vicinity. After perturbing the instance iteratively, a sample set is generated. An interpretable model with this sample is trained. Examples of these are linear regression, lasso, a decision tree with low depth. This tool is useful to understand if the model is right but for the wrong reasons \cite{ross2017right}. 

\vsp
\textbf{Explanations Vectors.} In 2010, Baehrens et al. \cite{baehrens2010explain} proposed a technique for explaining the local decision taken by a black-box in classification problems. They define an explanation vector as the derivative of the conditional probability function of the Bayes classifier. Having the explanation be a gradient vector captures a similar locality intuition to that of LIME. However, interpreting the coefficients on the gradient is difficult, particularly for confident predictions where gradient is close to zero \cite{ribeiro2016should}.

\vsp
\textbf{Counterfactuals.} The objective is to find an instance, close to the instance of interest, and yet that the prediction label is different. Therefore we want to find the smallest perturbation possible such that this condition is met. Otherwise, it would be hard to explain the difference in outcome values between instances that are radically apart. A counterfactual explanation of a prediction describes the minimal change in the feature values so that the prediction changes to a predefined outcome \cite{wachter2017counterfactual}. Defining a metric in the feature space is hard since there are several data types. Also, the Optimization Problem for finding \textit{the} smallest direction change is \textit{NP-hard} which means there is no known efficient algorithm that can solve it within a reasonable lapse of time. Finally, the solution must provide a real-world observable instance, i.e. feasible. 

\vsp
\textbf{Anchors.} In 2018, the same authors of LIME, Ribeiro et al. \cite{ribeiro2018anchors} proposed a newer method called Anchors, which extended the LIME using decision trees. They claim to have an efficient way to compute explanations for any black-box model with high-probability guarantees. In LIME, they used locally linear models to explain since they can capture the relative importance of features in an easy-to-understand manner. But, the local coverage is not entirely clear, especially over unseen regions. The method used is based on if-then rules. They define an anchor explanation as a set of rules. If the condition is met, we can disregard the rest of the features because they will not change the prediction significantly. 

\subsection{Contrastive Approach}
\label{contrastive}

Since humans tend to make comparisons between observations' predictions, Lipton \cite{lipton1990contrastive} states that explanations need to be contrastive by comparing a focal instance against a small reference group. Sometimes, it is desired to find a cluster of close by instances that share similar predictions. Other times, they are contrasted against similar instances but with different prediction outputs---these are counterfactual explanations. They seek slight changes---or perturbations---that the input could have had to obtain a different label.

\vsp
\textbf{Counterfactuals Naturally Observed.} To explain a focal instance, similar observed instances with different output label are seeked. For this types of methods the Optimization Problem previously described does not apply. The Contrastivity of this method is liked because we compare against real-observed instances, instead of artificially generated. Thus, for this types of methods the Optimization Problem previously described for finding the smallest change so that the prediction changes does not apply. The challenge for this method is that it is hard to define a notion of similarity. Moreover, because of the curse of dimensionality, the instances may be sparse. This challenges are further described in the second part of this work \cite{carrillo2020individualcase}.

\vsp
\textbf{Prototype and Criticism.} In 2016, Kim et al. \cite{kim2016examples} proposed an approach where we find a selection of representative instances from the data, called \textit{prototypes}. Then, we find instances that are not well represented by those prototypes, called \textit{criticisms}. These prototypes do not necessarily have to be actual data, but they could a combination of instances. There are several ways to find prototypes, whereas criticisms require a different methodology. In this article they give a framework to find prototypes, as well as criticisms. As for how to use it for individual explanations, first we need a way to understand the prototypes and criticism. Then, just take the closest prototypes or criticism to the instance of interest, and the same explanation should fit.

\vsp
\textbf{Justified Counterfactual Explanations} In 2019, Laungel et al. \cite{laugel2019dangers}, for classification problems, discussed the risk of making counterfactual explanations based on artifacts learned by the model, instead of actual knowledge. This problem arises for instances that lie in a subspace where there is no reliable information. For example, a point in the data where there are no nearby observed instances, or perhaps there are nearby instances, but all have different predicted labels. Or perhaps they have consistent labels but are incorrect. These three types of problems lead to the same result, explaining with unjustified counterfactuals. They propose a method to define the notion of justified counterfactuals as having sufficient and consistent correct predictions around the instance of interest.

\section{Model-Specific Methods}
\label{model_specific}
In this section, we describe briefly the most relevant and novel, local Model-Specific methods, which are enlisted in Table \ref{methods}. Most of the Model-Specific methods have focused on Neural Networks \cite{fong2017interpretable}. In Section \ref{machine} Machine Vision Models we reviewed, which use a particular type of Neural Network called Convolutional Neural Network. In Section \ref{general} we describe all the other methods for Neural Networks that are not necessarily focused on Machine Vision. Studies that do not use Neural Networks, such as gradient boosting tree-based prediction \cite{stojic2019explainable}, tend to use the explainability methods described in Section \ref{model_agnostic}, with a novel exception for Trees which is described in Section \ref{decision}. 

\subsection{Machine Vision Models}
\label{machine}
In this subsection we review methods for image classification. A common approach has been to find regions of an image that were particularly influential to the final classification. 

\vsp
A common approach is to use saliency maps. We will describe them briefly and review first the methods that use them. Saliency maps, also known as Sensitivity Maps, or Pixel Attribution Maps, use occlusion techniques or calculations with gradients to assign an ``importance'' value to individual pixels, which is meant to reflect their influence the final classification \cite{smilkov2017smoothgrad}.  They can be defined as the smallest region whose removal causes the classification score to drop significantly. By using these maps, studies \cite{dabkowski2017real} have been made to exhibit the flaws of the model. By perturbing an image slightly, even by one pixel \cite{bach2015pixel}, it has been shown that we can trick the algorithm into making a different prediction. Therefore, we cannot guarantee that we will not encounter such images in real data, showing that the model is not trustworthy. Saliency maps are usually rendered as Heat Maps \cite{dabkowski2017real}.

\vsp
\textbf{Masks.} In 2017, Fong and Vedaldi. \cite{fong2017interpretable} proposed a method similar to the paradigm of LIME. By perturbing an image slightly, they wish to learn an image perturbation mask that minimizes a class score. To do so, they focus on identifying which regions of an image are used by the black box to produce the output's value---in other words, finding maximally informative deletion regions. They explicitly edit the image, unlike other saliency techniques. 

\vsp
\textbf{Real-Time Saliency Maps.} In 2017, \cite{dabkowski2017real} developed a fast saliency detection method that can be applied to any differentiable image classifier. Previous approaches develop a saliency map by removing distinct patches of the image of interest, but this is computationally costly. Here, they train a model to predict such a map for any input image in a single feed-forward pass. According to the authors, the method results so fast compared to past approaches that can be used for Real-Time Image explanations. 

\vsp
\textbf{SmoothGrad.} In 2017, based on Saliency Maps, from Smilkov et al. \cite{smilkov2017smoothgrad}. The approach is to find those pixels that strongly influence the prediction by using gradients.  They mention that although the saliency maps sometimes make sense for humans, sometimes the explanation algorithm shows some ``important'' pixels that seem to be chosen randomly. They proposed SmoothGrad, which tries to reduce the noise of an image, so the sensitivity maps result trustworthy. Counterintuitively, the algorithm begins by adding noise. Given an image of interest, generate a sample by perturbing it, adding noise, and take the average importance of the pixels to produce the final sensitivity map.

\vsp
\textbf{Layer-wise Relevant Propagation.} In 2015, Bach et al. \cite{bach2015pixel} also proposed a method that focuses on finding the most determinant pixels for the model's output. The contribution of each pixel is appreciated in the shape of a heat map. They describe it as a toolset for deconstructing a nonlinear decision and fostering transparency for the user. To find the relevance of the pixels, they proposed two techniques, one uses Taylor decomposition, and the other one works similarly to backpropagation, which grants it of being efficient. 

\vsp
\textbf{Heat Maps.} In 2013, Zeiler and Fergus \cite{zeiler2014visualizing} published a highly relevant paper about heat maps. These highlight the most relevant parts of the input image to the neural networks' decision on a particular classification target. They use gradients and backpropagation to decide the relevance of the pixels.

\subsection{General Neural Networks}
\label{general}

In this section, we discuss methods to explain Neural Networks models. Some authors in their work make a distinction between the method being for Neural Networks or Deep Networks. Because we believe this line is still subtle, we treat both terms ambivalently, and make no distinction.   

\vsp
\textbf{Differentiable Models.} In 2017, Ross et at. \cite{ross2017right} explained that methods such as LIME or Saliency Maps are useful tools to identify if the model is right but for the wrong reasons. But after the model reveals its errors, it shows how to correct them. Also, they show that specifically for LIME, sometimes it explains mistakenly, or unfaithfully, when training and testing data differ from each other, in some sense. They propose a method that corrects this problem, with penalization over the gradients on the Neural Network. When annotations are given for those instances that are correct but for the wrong reasons, this method allows the classifier to learn alternate explanations. If the annotations are not given, they found a sample of equally valid explanations so that a human expert can decide the most reasonable explanation.

\vsp
\textbf{DeepLIFT.} In 2016, Shrikumar et al. \cite{shrikumar2016not} proposed a method called DeepLIFT, which stands for Deep Networks, while LIFT for Learning Important Features. It is a method for computing feature importance scores in a deep neural network. They claim that one of the main problems with gradient methods, such as the ones that generate saliency maps, originates from the popular ReLU activation functions that have gradient zero when they are not firing despite caring information. Instead, this method compares the activation of each neuron against its \textit{reference activation}. The reference activation is obtained from the activation of each neuron when applying a \textit{reference input}.

\vsp
\textbf{Taylor decomposition.} In 2017, Montavon et al. \cite{montavon2017explaining} broke the classification output into the contribution of its input elements. The method is called DeepTaylor decomposition. It can be used to assess the importance of the most relevant pixels in an image in which the explanation is shown as a heat map.

\vsp
\textbf{Integrated Gradients.} In 2017, Sundararajan et al. \cite{sundararajan2017axiomatic} proposed a method called Integrated Gradients. It is a simple method that can be implemented quickly to the Deep Network. It is also theoretically well sustained because the method satisfies two axioms: sensitivity and Implementation Invariance. Start by defining explicitly an image, which is the baseline, with the lowest prediction score, in the $n$th dimension, where $n$ is the size of the picture. The baseline image can be the all-black image, for example. Then, take an input of interest and define the segment of a line in the $n$-dimension that joins both images. We then calculate the path integral across this line. The visualization of integrated gradients can be seen as a heatmap. However, the heatmaps generated by integrated gradients are generally diffuse, and so difficult for humans understanding.

\vsp
\textbf{I-GOS.} In 2019, Qi et al. observed that \cite{qi2019visualizing}, the fact that heat maps do not correlate with the network may mislead the human. In other words, heat maps sometimes do not provide faithful explanations.  The common mask approach to finding the Heat Maps solves an optimization problem. Namely, to find the smallest and smoothest area that maximally decreases the output of a neural network. Still, this approach can be inefficient and get stuck in local minima. In this article, they proposed a method called I-GOS, which utilizes the integrated gradients to improve the mask optimization approach. The approach computes descent directions based on the integrated gradients instead of normal gradients.

\vsp
\textbf{Grad-cam.} In 2017, Selvaraju et al. \cite{selvaraju2017grad} proposed to use a heat map technique, which has become quite popular. They called it Gradient-weighted Class Activation Mapping, or simply Grad-cam. It highlights crucial regions in the image for predicting the class. When a seemingly unreasonable prediction was given, the algorithm gives a reasonable explanation for why this happened. 

\subsection{Decision Trees Methods}
\label{decision}

To our knowledge few interpretability methods for other than Neural Networks have been proposed. To explain these, they use Model-Agnostic methods. \cite{stojic2019explainable}. Still, we found a novel article for models based on trees which we found interesting. 

\vsp
\textbf{TreeExplainer.} In 2020, Lundberg et al. \cite{lundberg2020local} exposed that models based on trees are the most popular nonlinear models nowadays. This group of models contains random forest, gradient boosted trees, and other tree-based models. They are known for being accurate and interpretable. The latter, in the sense that we can understand which features were used to make the prediction. Despite this, little attention has been laid on explaining individual predictions. The current local explanation for this type of model include (i) reporting the decision path, (ii) assign the contribution for each feature, and (iii) applying a model-agnostic approach. Each of these alternatives has limitations. For (i), it is unhelpful when the model uses multiple trees for the final prediction. For (ii), the explanations are likely to be biased. For the (iii), they can be slow and suffer from sampling variability. They propose a method that does not have any of the stated deficiencies. The model is called TreeExplainer and is based on the SHAP.  

\section{Conclusion}
\label{conclusion}

In this survey, we reviewed the relevant and novel approaches that form the state-of-the-art methods which address the problem of explaining individual instances in ML. Two decades ago, this topic was both unexplored as well as irrelevant. But as algorithmic decision-making has spread increasingly to many relevant societal contexts, and the trend of the most successful algorithms has gradually moved towards the use of highly complex models, it has become increasingly desirable to explain the model's predictions.

\vsp
The interpretability methods vary on explanation types. Some use natural language, other visualizations of learned representations or models, and some are example-based, comparing focal instances against reference groups. We classified according to Model Generability. It divides the methods by a Model-Agnostic approach and a Model Specific approach. The difference is that for the former, the interpretability methods can be applied to any type of ML model. In contrast, for the latter, the methods can be applied to only a particular class of models. We further subclassified each taxonomy. The Model Agnostic Approach by Perturbation-Based Approach---e.g., SHAP, Sensitive Analysis, or LIME---and the Contrast Approach---e.g., Counterfactuals or Prototype and Criticism methods. 

\vsp
As for the Model-Specific Methods, these were subclassified by Convolutional Neural Networks (Computational Vision), General Neural Networks, and Decision Tree methods. We found that most Model-Specific Methods are designed for Neural Networks---whether for Computational Vision or a General Purpose---and only a few explain other types of complex algorithms, such as random forests and gradient boosting regressors. However, recently this family of models has outperformed Neural Networks for relevant societal context. Thus, in the upcoming years, we believe a natural shift will occur towards proposing methods for complex Decision Tree based models instead. 

\vsp
We aim that this work can be useful for practitioners seeking to implement explainability layers into complex statistical models that influence decisions in real-world contexts.

\begin{multicols}{2}
\bibliographystyle{unsrt}
\bibliography{main}
\end{multicols}

\end{document}